\documentclass[conference]{IEEEtran}
\IEEEoverridecommandlockouts
\usepackage{cite}
\usepackage{amsmath,amssymb,amsfonts}
\usepackage{algorithmic}
\usepackage{graphicx}
\usepackage{textcomp}
\usepackage{url}
\usepackage{multirow}
\usepackage{xcolor}
\usepackage{stfloats}
\usepackage{float}
\def\BibTeX{{\rm B\kern-.05em{\sc i\kern-.025em b}\kern-.08em
    T\kern-.1667em\lower.7ex\hbox{E}\kern-.125emX}}
\begin{document}

\title{Explainable Deep Learning in Medical Imaging: Brain Tumor and Pneumonia Detection}

\author{\IEEEauthorblockN{Sai Teja Erukude}
\IEEEauthorblockA{\textit{Department of Computer Science} \\
\textit{Kansas State University}\\
Manhattan, USA \\
erukude.saiteja@gmail.com}
\and
\IEEEauthorblockN{Viswa Chaitanya Marella}
\IEEEauthorblockA{\textit{College of Business Administration} \\
\textit{Kansas State University}\\
Manhattan, USA \\
viswachaitanyamarella@gmail.com}
\and
\IEEEauthorblockN{Suhasnadh Reddy Veluru}
\IEEEauthorblockA{\textit{College of Business Administration} \\
\textit{Kansas State University}\\
Manhattan, USA \\
suhasnadhreddyveluru@gmail.com}
}

\maketitle

\begin{abstract}
Deep Learning (DL) holds enormous potential for improving medical imaging diagnostics, yet the lack of interpretability in most models hampers clinical trust and adoption. This paper presents an explainable deep learning framework for detecting brain tumors in MRI scans and pneumonia in chest X-ray images using two leading Convolutional Neural Networks, ResNet50 and DenseNet121. These models were trained on publicly available Kaggle datasets comprising 7,023 brain MRI images and 5,863 chest X-ray images, achieving high classification performance. DenseNet121 consistently outperformed ResNet50 with 94.3 percent vs. 92.5 percent accuracy for brain tumors and 89.1 percent vs. 84.4 percent accuracy for pneumonia. For better explainability, Gradient-weighted Class Activation Mapping (Grad-CAM) was integrated to create heatmap visualizations superimposed on the test images, indicating the most influential image regions in the decision-making process. Interestingly, while both models produced accurate results, Grad-CAM showed that DenseNet121 consistently focused on core pathological regions, whereas ResNet50 sometimes scattered attention to peripheral or non-pathological areas. Combining deep learning and explainable AI offers a promising path toward reliable, interpretable, and clinically useful diagnostic tools.
\end{abstract}


\section{Introduction}\label{introduction}

Recent developments in Deep Learning have led to significant progress in medical imaging, with models achieving expert-level performance in various diagnostic tasks \cite{Yasaka2018}. Convolutional Neural Networks (CNNs), in particular,  have been successfully applied to tasks such as tumor detection in MRI scans, pneumonia detection in chest X-rays, and diabetic retinopathy screening in retinal images. These models can learn underlying patterns from imaging data and display high sensitivity to abnormalities in radiographs and MRI scans \cite{M2024}.

However, a significant obstacle to the practical deployment of these models is their poor interpretability and the perceived ``black-box" nature of their decision-making. Clinicians, for instance, have a difficult time trusting the output of AI systems without understanding how they made the prediction \cite{info16010053}. Traditional CNNs have millions of parameters and layers that are non-interpretable, and it is difficult to explain how a specific prediction was made. Moreover, hidden background noise or patterns irrelevant to the subject of interest also influence the decision-making process of deep neural networks. This has been proved by evaluating the models on a part of the background and analyzing the images in the frequency domain \cite{computers2024, erukude2024identifying}. Thus, explainability in AI (XAI) is especially important in clinical settings, as it ensures that recommendations are transparent and in line with clinical reasoning. By providing human-interpretable justifications, XAI methods can facilitate physician trust and improve the uptake of AI in diagnostic applications \cite{make6040111}.

This paper contributes to the need for explainable deep learning in two important medical diagnostic tasks: brain tumor classification using MRI scans and pneumonia detection from chest X-rays. Two widely used and popular CNN architectures were employed for image recognition tasks, ResNet50 and DenseNet121. ResNet50 is a 50-layer deep residual network that uses a skip connection to reduce a block of layers into a single layer \cite{he2015deepresiduallearningimage}. DenseNet121 is a deep convolutional network that builds on ResNet, connecting every layer to every subsequent layer, allowing for improved feature reuse and better gradient flow \cite{huang2018denselyconnectedconvolutionalnetworks}. DenseNet models have been frequently used in medical image classification tasks (for example, CheXNet for pneumonia) \cite{rajpurkar2017chexnetradiologistlevelpneumoniadetection}. These models were separately trained on two curated publicly available Kaggle datasets, the brain tumor MRI dataset \cite{mrikaggle} and chest x-ray dataset \cite{chestxraykaggle}, achieving robust performance across both tasks.

To tackle the interpretability gap, an explainability technique called Gradient-weighted Class Activation Mapping (Grad-CAM) was employed. It uses gradient information back-propagated into the last convoluted layer of a CNN, yielding a heatmap highlighting the most influential regions in the input image for the model's prediction \cite{Selvaraju_2019}. This approach allows us to visually explain each class without modifying the model's architecture and enhances the model's transparency. Clinicians can now visually verify that the AI is focusing on medically relevant parts of the image (e.g., areas of tumor tissue on an MRI or infiltrates in a lung X-ray) or is distracted by spurious features.

The key contributions of this study are twofold: firstly, a systematic comparison between ResNet50 and DenseNet121 on brain MRI and chest X-ray datasets is provided, which illustrates their respective strengths in medical image classifications. Secondly, it is demonstrated that Grad-CAM is a communicable, human-interpretable explanation that is consistent with expert understanding. It is believed that the combination of high-performing models with explainable outputs will be a foundation to increase trust, improve clinical acceptance, and ultimately, AI systems in health care will be platforms for real-world deployments.

\section{Literature Review}

Over the past decade, deep learning has fundamentally changed how medical images are analyzed and interpreted \cite{Yasaka2018, M2024, shen2017deep}. Unlike past computer-aided diagnosis systems that used hand-crafted features, CNNs can learn rich feature representations from raw imaging data. Litjens et al. (2017) provided a comprehensive survey highlighting over 80 applications of deep learning in medical imaging, including many advancements in accuracy that were superior to traditional post-analytic methods \cite{Litjens2017Survey}. There have been a few examples of CNN models approaching expert-level classification of skin lesions and retinal disease from fundus images. In radiology, models are being developed that detect pathologies in chest radiographs, breast cancer on mammograms, etc. Another appeal of the deep learning paradigm is accuracy. Once the algorithms are trained, the inference is much quicker, reducing time for examination and follow-up in situations where radiologists are flooded with images. 

Among CNN architectures, ResNet and DenseNet have emerged as two of the most influential designs in recent years. ResNet (Residual Network) employs residual learning with skip connections, creating a direct pathway for gradients to the input and helping combat vanishing gradients \cite{he2015deepresiduallearningimage}. This allows successful training of very deep networks such as ResNet50 and ResNet152. DenseNet (Densely Connected Network) takes a different approach by introducing dense connectivity, where each layer is connected feed-forward to every other layer. Specifically, the feature maps of all preceding layers are concatenated and provided to each subsequent layer as input \cite{huang2018denselyconnectedconvolutionalnetworks}.
For instance, DenseNet architectures such as DenseNet121 have been found to outperform traditional CNNs and even ResNets in several medical tasks, including pneumonia detection from chest X-rays \cite{rajpurkar2017chexnetradiologistlevelpneumoniadetection, Zhou2022DenseNet}, due to their efficient parameter usage and ability to learn fine-grained patterns.

However, despite their impressive predictive capabilities, both ResNet and DenseNet suffer from a lack of interpretability, which is a crucial factor in the healthcare sector. In response to this shortcoming,  explainable AI (XAI) methods such as Gradient-weighted Class Activation Mapping (Grad-CAM) have been developed to provide visual explanations for CNN decisions \cite{Selvaraju_2019}. Grad-CAM uses backpropagation to compute the score gradient for the target class relative to the feature maps of the final convolutional layer, and it averages these gradients. Grad-CAM then weighs the feature maps using gradients. The result is a heatmap that can be up-sampled and overlaid on the original image to visualize important regions that the model included in its decision \cite{Selvaraju_2019}.

In a recent study, Musthafa et al. (2024) implemented Grad-CAM in conjunction with ResNet50 to detect brain tumors. The heatmaps indicated that the major emphasis was on the tumor areas of the MRI scans, which corresponded with the radiologist's expectations \cite{M2024, make6040111}. In conclusion, the existing literature suggests that XAI techniques like Grad-CAM will be a key asset for clinical AI applications because they give a reasonable way to ``open" the black box and present the model's thinking in a way humans understand. This study builds on this body of work by integrating Grad-CAM into both ResNet50 and DenseNet121 models for brain tumor and pneumonia detection, offering not only high accuracy but also intuitive, human-understandable visual explanations.

\section{Methodology}

\subsection{Data}
The proposed approach was assessed on two public medical imaging datasets available on Kaggle: a brain MRI dataset for tumors classified as either glioma tumor, meningioma tumor, pituitary tumor, or no tumor (healthy brain) \cite{mrikaggle}, and a chest X-ray dataset for identifying pneumonia \cite{chestxraykaggle}.

\subsubsection{Brain MRI Dataset}
\begin{figure}[ht]
    \centering
    \includegraphics[width=3.3in]{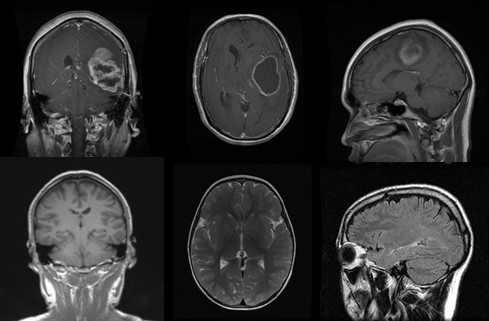}
    \caption{Sample images from the Brain MRI dataset showing tumor, and no tumor cases.}
    \label{fig_brain_tumor}
\end{figure}

The Brain MRI Dataset comprises T1-weighted contrast-enhanced MRI images of human brains. There are more than 7000 MRI slices, labeled by radiologists, making this dataset more reliable in terms of ground truth. The dataset initially contained four classes (glioma tumor, meningioma tumor, pituitary tumor, and no tumor), which were then converted into just two classes (tumor and no tumor), as shown in the Figure \ref{fig_brain_tumor}.

\subsubsection{Chest X-ray Dataset}
\begin{figure}[ht]
    \centering
    \includegraphics[width=3.3in]{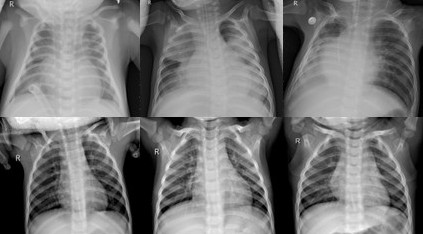}
    \caption{Sample images from the Chest X-ray dataset showing Pneumonia and normal cases.}
    \label{fig_chest_xrays}
\end{figure}

The Chest X-ray dataset comprises about 5863 frontal chest X-ray images of pediatric patients with two categories: Pneumonia (positive cases with radiological evidence of pneumonia) and Normal (healthy lungs). These images were initially taken as part of clinical routine processes at Guangzhou Women and Children's Medical Center and have been screened for quality. This dataset gives a binary classification problem for detecting pneumonia, which typically presents as opacities in the lung fields on the X-ray image, as depicted in the Figure \ref{fig_chest_xrays}. Pneumonia may be due to bacterial or viral infections.

Each dataset was divided into a standardized train-validation-test split of 70\% training, 15\% validation, and 15\% testing (or hold-out) sets. Moreover, on-the-fly data augmentation was performed to enhance training diversity and model robustness. Images are rescaled to normalize pixel values. Random rotations (up to 15°), along with width and height shifts of up to 10\%, and zoom variations of up to 10\%. simulate real-world variations in object positioning and size. Horizontal flipping helps the model learn from mirrored images, while the ``reflect" fill mode preserves edge continuity during transformations.

\subsection{CNN Architectures}

Two CNN architectures, ResNet50 and DenseNet121, were trained on the MRI and X-ray tasks (four model instances in total). These models were initialized with ImageNet pre-trained weights (a standard practice in transfer learning), which were an adequate starting point for this study, given that features from ImageNet essentially transfer well onto medical images for low-level patterns. 

\begin{figure*}[hbt!]
    \centering
    \includegraphics[width=5.5in]{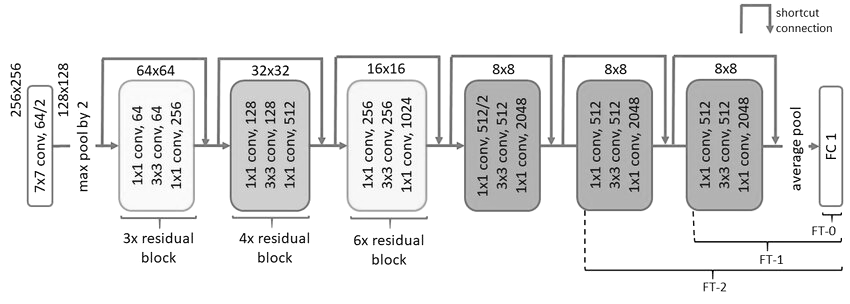}
    \caption{High-level architecture of the ResNet50 deep neural network used for both brain MRI and Pneumonia classification tasks.}
    \label{fig_resnet}
\end{figure*}

ResNet50 has 50 layers that consist of 16 residual blocks (with identity shortcut connections) and uses batch normalization and ReLU activation. Its depth and residual design, shown in the Figure \ref{fig_resnet}, make it capable of learning complex visual features. DenseNet121 uses four dense block sections and 121 layers with a growth rate of 32 per layer. Each dense block is transitioned to a pooling and convolution layer. DenseNet121 concatenates the features, meaning the final classifier also sees feature maps from numerous levels of abstraction. Despite having a more significant number of layers, DenseNet121 will have a similar number of parameters as ResNet50, due to its narrower layers and feature reuse strategy.

\begin{figure*}[hbt!]
    \centering
    \includegraphics[width=5.5in]{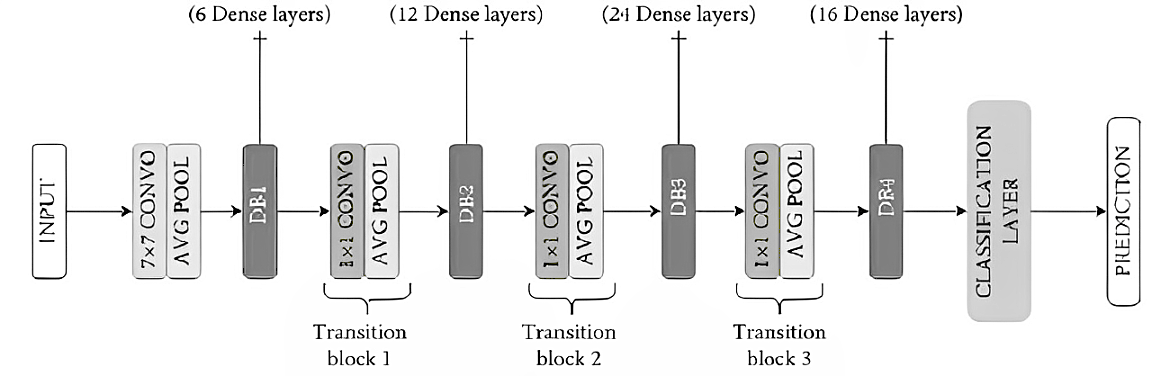}
    \caption{High-level architecture of DenseNet121 model illustrating dense connectivity and feature reuse across layers.}
    \label{fig_densnet}
\end{figure*}

\subsection{Training Setup}

All four models (MRI-ResNet, MRI-DenseNet, Pneumonia-ResNet, and Pneumonia-DenseNet) were trained using the Adam optimizer that began training with a learning rate of 0.0001. Adam was chosen because it adapts learning rates on a per-parameter basis per evaluation. The learning rate was scheduled to be reduced on a plateau: if the validation loss did not improve for five epochs, then the learning rate would be reduced by a factor of 0.2, using the `ReduceLRonPlateau' callback. The batch size was 32 and was trained for 50 epochs, with early stopping in case of convergence. Specifically, an `EarlyStopping' callback was used that monitored validation loss and had a patience of ten epochs; if the validation loss did not decrease over ten epochs, it would stop training to avoid overfitting. Additionally, `ModelCheckpoint' was employed to save the best model weights (using validation loss as a model quality metric) and return them to the best model weights after unconditional training.

\subsection{Evaluation Metrics}

The trained models were assessed on the hold-out testing set using various metrics: Accuracy, Area Under the Receiver Operating Curve (AUC), F1 Score, and Average Precision (AP). 

\begin{itemize}
    \item Accuracy: The total number of correct predictions is a direct reflection of overall model performance. Considering accuracy as the only metric can be misleading when dealing with unbalanced datasets. Hence, it is important to account for other metrics to provide a holistic view.
    
    \item AUC: Measures the model's ability to discriminate between classes across all classification thresholds; an AUC of 1.0 indicates perfect discrimination, and 0.5 indicates chance-level discrimination.
    
    \item F1 Score: Harmonic mean between precision and recall was calculated to provide an understanding of the model's measure of false positives in false negative space
    
    \item Average Precision: AP is the area under the precision-recall curve. This provides more weight to how the positive class performed in class-imbalanced scenarios.
\end{itemize}

These various metrics help us understand not only the number of correct predictions (accuracy) but also how robust the model is to threshold choice (AUC, AP) and how it balances sensitivity vs specificity (F1). Confusion matrices were also recorded for each model during testing to assess which classes are most confused with each other.

\section{Implementation and Results}

Both models achieved strong results on their respective tasks, with DenseNet121 outperforming ResNet50 consistently. Table \ref{tab_comparison} summarizes the test performance of ResNet50 vs DenseNet121 on brain MRI and chest X-ray datasets. 

\begin{table*}[htbp]
\caption{Comparison of ResNet50 and DenseNet121 Performance on Brain MRI Tumor Classification and Chest X-ray Pneumonia Detection.}
\centering
\begin{tabular}{|l|l|c|c|c|c|}
\hline
\textbf{Dataset} & \textbf{Model} & \textbf{Accuracy (\%)} & \textbf{AUC} & \textbf{F1-Score} & \textbf{Avg Precision} \\
\hline
\multirow{2}{*}{Brain MRI Tumor} & ResNet50 & 92.5 & 0.95 & 0.87 & 0.82 \\
                                   & DenseNet121 & \textbf{94.3} & \textbf{0.99} & \textbf{0.91} & \textbf{0.88} \\
\hline
\multirow{2}{*}{Chest X-ray Pneumonia} & ResNet50 & 84.4 & 0.95 & 0.89 & 0.82 \\
                                         & DenseNet121 & \textbf{89.1} & \textbf{0.98} & \textbf{0.93} & \textbf{0.88} \\
\hline
\end{tabular}
\label{tab_comparison}
\end{table*}

\begin{figure*}[hbt!]
    \centering
    \includegraphics[width=6.5in]{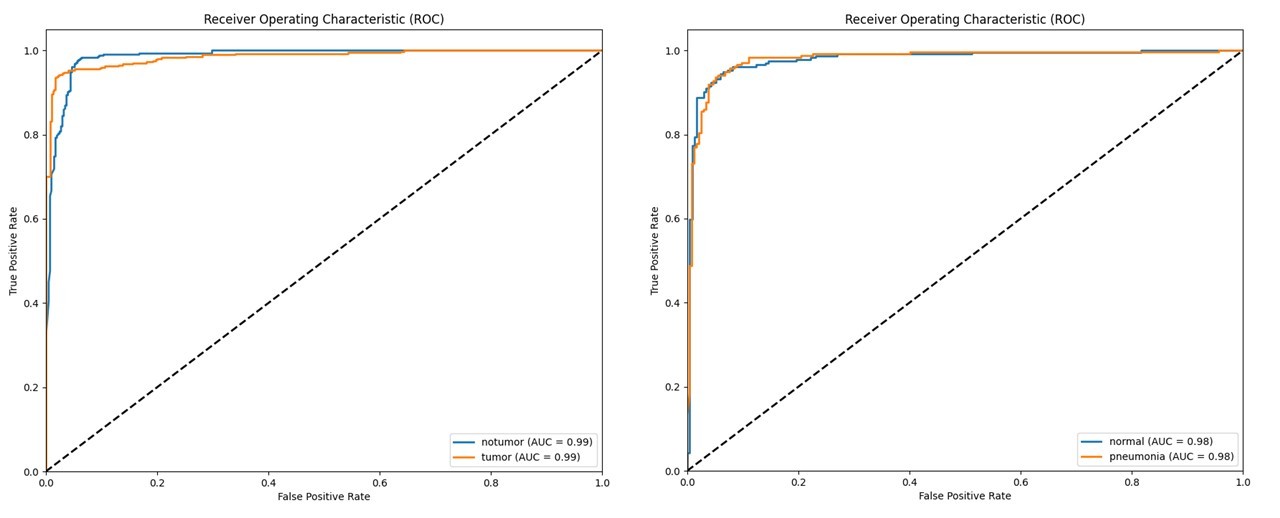}
    \caption{ROC curves illustrating the performance of DenseNet models on brain tumor (left) and pneumonia (right) classification tasks.}
    \label{fig_densenet_roc}
\end{figure*}

In the brain tumor classification task, ResNet50 achieved an accuracy of 92.5\%, while DenseNet121 attained an accuracy of 94.3\%. Although both models performed well with accuracy results, DenseNet121 provides a more vigorous precision and recall balance, where it achieved an F1-score of 0.91, and the ResNet50 score was around 0.87. The AUC for tumor classes showed a similar relationship with DenseNet121, slightly outperforming ResNet50 (around 0.99 compared to 0.95) as shown in the figure \ref{fig_densenet_roc}. However, both AUC scores were exceptionally high for distinguishing no tumor vs tumor, with DenseNet121 nearly flawless in the "no tumor" AUC in detecting the absence of a tumor. 

DenseNet121 showed a much more significant performance gap in the pneumonia presence detection task than ResNet50. ResNet50 achieved an accuracy of 84.4\% on the test X-rays, while DenseNet121 attained an accuracy of 89.1\%. The AUC-ROC for pneumonia vs regular class was 0.95 for ResNet50 and 0.98 for DenseNet121; in practice, at the optimal threshold, the DenseNet121 had higher sensitivity (recall) for pneumonia with only a slight compromise in specificity compared to ResNet50. The F1-score for ResNet50 was 0.81, while the DenseNet121 performed better with a 0.86 F1. 

DenseNet121 has a better balance of precision-recall curve and improved sensitivity to fine-grained textural infection patterns in lungs and a possible regularizing effect through feature reuse, which is beneficial considering the dataset size. These results are consistent with other studies showing that DenseNet-based models perform well or excel in medical image classification tasks, including chest X-rays \cite{rajpurkar2017chexnetradiologistlevelpneumoniadetection}.

Results confirm that DenseNet121 has an edge over ResNet50 in both scenarios. DenseNet's feature reuse is believed to be valuable in achieving the state-of-the-art performance. With the pneumonia task, the difference may be partially due to the smaller dataset; DenseNet's stronger gradients and feature reuse may better allow it to learn from limited data. ResNet50 was not too far behind, though, and provided good performance, indicating both architectures are effective for these medical classification tasks.

\section{Grad-CAM Explainability Results}

\begin{figure}[H]
    \centering
    \includegraphics[width=2.5in]{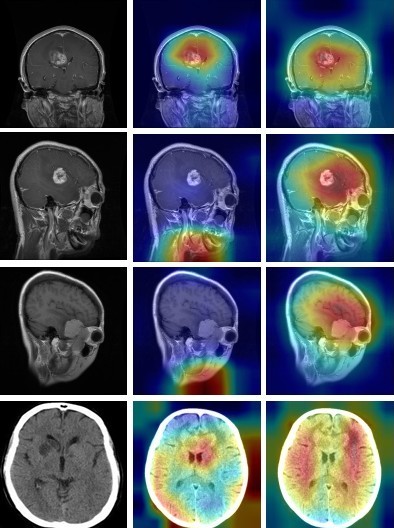}
    \caption{Grad-CAM visualizations for Brain MRI: (Left) original MRI image, (Middle) ResNet50 heatmap, (Right) DenseNet121 heatmap. Note how DenseNet121 focuses more precisely on the tumor region, while ResNet50 shows broader, sometimes scattered attention.}
    \label{fig_mri_cam}
\end{figure}

\begin{figure}[H]
    \centering
    \includegraphics[width=2.5in]{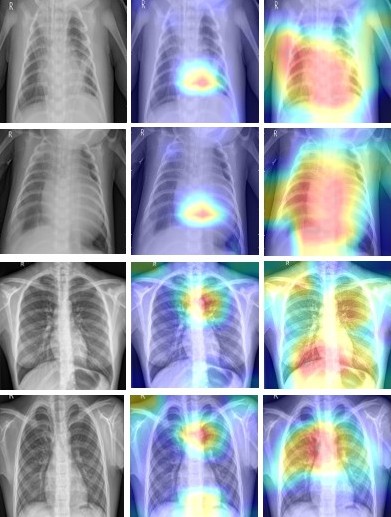}
    \caption{Grad-CAM visualizations for Chest X-ray: (Left) original X-ray image, (Middle) ResNet50 heatmap, (Right) DenseNet121 heatmap. DenseNet121 localizes attention to lung regions with opacities, while ResNet50 occasionally highlights unrelated areas such as the heart or ribs.}
    \label{fig_xray_cam}
\end{figure}

To make the model's decisions interpretable, Grad-CAM was employed on the entire test dataset for each task. Grad-CAM uses gradient information back-propagated into the last convoluted layer and creates a heatmap indicating the regions of the image that contributed the most to the predicted class. These heatmaps are then overlaid onto the original image, visualizing the model's attention. The representative examples of these Grad-CAM visualizations for the brain MRI and chest X-ray tasks are illustrated in Figure \ref{fig_mri_cam} and Figure \ref{fig_xray_cam}, respectively. The figures consist of three columns: (1) the original input image, (2) ResNet50 heatmap, and (3) DenseNet121 heatmap. These visualizations illustrate the different attention behavior between the two architectures.

For brain tumor detection (Figure \ref{fig_mri_cam}), both ResNet50 and DenseNet121 generated localized heatmaps mostly relevant to tumor areas. However, it was observed that DenseNet121 consistently highlighted the core tumor area, which would be very defined, and nearer to the contrast-enhancing or mass-like lesions apparent in MRI. In contrast, ResNet50, while correctly identifying the presence of a tumor, distributed attention on areas that were more diffuse (undefined tumor), and sometimes it developed emphasis on hyper-peripheral areas, and areas that didn't include the tumor at all, such as the skull or adjacent normal tissue. This suggests that while ResNet50 can capture broad features related to pathology, DenseNet121’s dense connectivity may have focused more sharply on the most discriminative features, avoiding distraction from irrelevant parts of the image.

In chest X-ray pneumonia detection (Figure \ref{fig_xray_cam}), similar differences were noticed. DenseNet121 heatmaps concentrated on lung fields with opacities, especially in the lower and middle lobes, where pneumonia is usually present in most cases. ResNet50, on the other hand, often displayed activation around the central region or silhouette of the heart, with some models even producing attention around the diaphragm or rib structures. While both models returned accurate predictions, DenseNet's concentration was more common in radiological signs of pneumonia, suggesting that its feature extraction layers prioritize more relevant and meaningful pathology. This behavior suggests that DenseNet121 might also generalize better to unseen pneumonia patterns by concentrating on specific lung textures and avoiding being biased to general images or anatomical landmarks.

Overall, the Grad-CAM analysis reveals meaningful differences in the interpretability profiles of the two models. While both architectures achieved a comparable high predictive performance, DenseNet121 consistently produced more clinically aligned explanations by focusing on true pathological regions in the image. Whereas ResNet50 sometimes exhibits attention scatter, which could introduce potential risks if over-interpreted by clinicians. These findings reinforce that when AI models are assessed for the health-care sector, accuracy metrics should always be paired with explainability tools; the model not only should be accurate but also confirm it is interpreting for the right reasons. In a clinical deployment, such differences could impact trust and usability: a model that points to the tumor or lung opacity directly will be more easily validated and trusted by human experts than one that highlights less relevant areas.

\section{Conclusion}

This study presented an explainable deep learning framework for brain tumor classification using MRI scans and pneumonia classification using chest X-rays. In particular, two of the most popular CNNs, ResNet50 and DenseNet121, were leveraged. Results showed that both models provided strong predictive performance, DenseNet121 consistently outperformed ResNet50 in accuracy and AUC scores, but also in the quality of visual explanations. Grad-CAM analysis revealed that DenseNet121 precisely focused on the key pathological regions, whereas ResNet50 occasionally attended to less relevant areas. Illustrating the importance of combining numerical evaluation with qualitative interpretability assessments. The results highlight that explainable AI is not an abstract appendage but a critical consideration for clinical implementation, as it enables health professionals to review and validate model decisions. Combining Grad-CAM visuals increases collaborative potential to provide accountable, reliable, transparent systems and actions within diagnostic workflows. Overall, this work demonstrates that combining high-performing architectures like DenseNet121 with explainability tools offers a promising path toward real-world, trustworthy AI in medical imaging. Future efforts will focus on expanding to new diseases, testing advanced XAI techniques, and deploying these models in clinical environments to fully realize their potential in improving patient care. Looking ahead, there are several promising directions to extend this research. Broadening to more diseases and imaging types, exploring advanced explainability methods (Grad-CAM++, Score-CAM, or new types of integrated gradients), and conducting clinician-centered user studies are some of the efforts that can help deploy explainable deep learning systems in diagnostic workflows and improve patient outcomes.

\section*{ACKNOWLEDGMENT}

The authors thank the contributors of the public datasets used in this study. The complete Python code has been made available at: \url{https://github.com/SaiTeja-Erukude/xai-in-medical-imaging}.

\bibliographystyle{IEEEtran}
\bibliography{main}

\end{document}